\documentclass{ceurart}

\usepackage[frozencache=true,cachedir=minted-cache]{minted}
\usepackage{booktabs}
\usepackage{tikz}
\usetikzlibrary{shapes.geometric, arrows, calc}

\tikzset{
    block/.style = {
        rectangle, 
        draw, 
        text width=5em, 
        text centered, 
        minimum height=4em,
        rounded corners
    },
    line/.style = {
        draw, 
        -latex',
    },
}
\begin{document}

\copyrightyear{2023}
\copyrightclause{Copyright for this paper by its authors.
  Use permitted under Creative Commons License Attribution 4.0
  International (CC BY 4.0).}

\conference{The First International Workshop on the Future of No-Code Digital Apprentices,
  August 19--25, 2023, Macao, S.A.R}

\title{Semantic Parsing for Complex Data Retrieval: Targeting Query Plans vs. SQL for No-Code Access to Relational Databases}

\author[1]{Ben Eyal}[%
orcid=0000-0002-1160-8973,
email=bene@post.bgu.ac.il,
]
\address[1]{Ben-Gurion University of the Negev, Be'er Sheva, Israel}

\author[1]{Amir Bachar}[%
email=amirbac@post.bgu.ac.il,
]

\author{Ophir Haroche}[%
email=ophirbh@gmail.com,
]

\author[1]{Michael Elhadad}[%
orcid=0000-0002-5629-2351,
email=elhadad@bgu.ac.il,
]

\begin{abstract}
    Large Language Models (LLMs) have spurred progress in text-to-SQL, the task of generating SQL queries from natural language questions based on a given database schema. Despite the declarative nature of SQL, it continues to be a complex programming language. In this paper, we investigate the potential of an alternative query language with simpler syntax and modular specification of complex queries. The purpose is to create a query language that can be learned more easily by modern neural semantic parsing architectures while also enabling non-programmers to better assess the validity of the query plans produced by an interactive query plan assistant.

    The proposed alternative query language is called Query Plan Language (QPL). It is designed to be modular and can be translated into a restricted form of SQL Common Table Expressions (CTEs). The aim of QPL is to make complex data retrieval accessible to non-programmers by allowing users to express their questions in natural language while also providing an easier-to-verify target language. The paper demonstrates how neural LLMs can benefit from QPL's modularity to generate complex query plans in a compositional manner. This involves a question decomposition strategy and a planning stage.

    We conduct experiments on a version of the Spider text-to-SQL dataset that has been converted to QPL. The hierarchical structure of QPL programs enables us to measure query complexity naturally. Based on this assessment, we identify the low accuracy of existing text-to-SQL systems on complex compositional queries. We present ways to address the challenge of complex queries in an iterative, user-controlled manner, using fine-tuned LLMs and a variety of prompting strategies in a compositional manner.

\end{abstract}


\maketitle

\section{Introduction}

Relational databases constitute the most widespread form of structured data storage. Querying and exploring complex relational data stores necessitate programming skills and domain-specific knowledge of the data. However, recent advancements in Natural Language Processing (NLP) have made it feasible to formulate questions about relational data in natural language, convert the questions into SQL, and present the results.

To facilitate the training of modern neural models for the text-to-SQL task, several large datasets have been created \cite{textosql-survey-2022}. These datasets consist of examples of (question, SQL query) pairs on a specific database schema, such as the MIMICSQL \cite{mimicsql-2020} and SEDE \cite{sede-2021} datasets for complex queries, or simple queries on a single table at a time for many tables, such as the WIKISQL  dataset \cite{wikisql-2017}. Notably, the \textit{Spider} dataset \cite{spider2018} encompasses multiple database schemas with complex queries, and it is employed to assess the generalization capabilities of text-to-SQL models to unseen schemas on complex queries.

When querying relational databases, SQL has become a ubiquitous tool. Consequently, it is natural to focus on generating SQL queries when translating natural language questions. The field of text-to-SQL has benefited from recent advancements in text-to-code \cite{CodeBERT-2020}. It has also been approached as a translation task from natural language to a structured formalism, such as lambda-calculus or logic formula, using attention-based sequence-to-sequence neural architectures \cite{dong-lapata-2016-language}. Previous research in this semantic parsing perspective has shown that the choice of the target language impacts a model's ability to learn to parse text into an accurate semantic representation. For instance, \citet{guo-etal-2020-benchmarking-meaning} compared the performance of various architectures on three question-answering datasets with targets converted to Prolog, Lambda Calculus, FunQL, and SQL. They discovered that the same architectures produce lower accuracy (up to a 10\% difference) when generating SQL, indicating that SQL is a challenging target language for neural models. Similarly, \citet{recogs-2023} determined that syntactic aspects of the target language in semantic parsing, such as redundant tokens and variable naming and position indices, have an effect on how well a given architecture can generalize on a dataset.

\begin{figure*}[t]
\raggedright
\textbf{Question}: What is the total ticket expense of the visitors whose membership level is 1?

\textbf{\\Original SQL query:} 
    \begin{minted}{sql}
    SELECT SUM(t2.Total_spent) 
    FROM museum_visit.visitor AS t1 JOIN 
         museum_visit.visit AS t2 
         ON t1.id = t2.visitor_id
    WHERE t1.Level_of_membership = 1
    \end{minted}

\textbf{\\QPL Plan:}
    \begin{minted}{text}
[
  [
    Scan Table [visitor] Predicate [visitor.Level_of_membership = 1] Output [ID],
    Scan Table [visit] Output [visitor_ID, Total_spent]
  ] Into: Join Predicate [visitor.ID = visit.visitor_ID] Output [visit.Total_spent]
] Into: Aggregate Output [SUM(visit.Total_spent)]
\end{minted}
    \caption{Sample Question, SQL Query and the corresponding QPL Query Plan. Every node in the QPL plan is executable and outputs a stream of tuples.} 
    \label{fig:plan}
\end{figure*}

A variety of innovative techniques have been applied to the task, leading to impressive overall accuracy gains. For example, the \textit{Spider} leader-board reports accuracy rates jumping from about 40\% in 2020 to about 85\% recently. Yet, our analysis indicates that when dealing with \textit{complex queries}, even the most advanced models fail to produce accurate SQL queries (on complex queries, accuracy falls to about 40\%). The problem of complex queries is acute when we consider the context of digital assistants for non-programming users: LLMs, and text-to-code models specifically, are well-known to produce incorrect results with confidence, and complex SQL queries will be hard to detect, decipher and validate by users.

In this paper, we address the challenge of dealing with complex data retrieval contexts. Our strategy consists in defining an intermediary compositional query language, which is easier to learn using LLMs and easier to explain to non-expert users. The query language reflects the semantics of complex SQL queries and can be translated directly into executable modular SQL programs. 

In the rest of the paper, we present recent progress in text-to-SQL models. We then present the Query Plan Language (QPL) we have designed and the conversion procedure we have implemented to translate the existing large-scale \textit{Spider} dataset into QPL. Analysis of the QPL data allows us to determine the success rate of leading text-to-SQL data on the more complex subset of the dataset.  We then present multiple strategies that exploit the compositional nature of QPL to train models capable of predicting complex QPL query plans from natural language questions.

\section{Previous Work}

\subsection{Architectures for text-to-SQL}

Text-to-SQL parsing consists of mapping a question $Q = (x_1, \ldots, x_n)$
and a database schema $S = [table_1(col^1_1 \ldots col^1_{c_1}), \ldots, table_T(col^T_1 \ldots col^T_{c_T})]$ into a valid SQL query $Y = (y_1, \ldots, y_q)$. Performance metrics include exact match (where the predicted query is compared to the expected one according to the overall SQL structure and within each field token by token) and execution match (where the predicted query is executed on a database and results are compared).

Since the work of \citet{dong-lapata-2016-language}, leading text-to-SQL models have adopted attention-based sequence to sequence architectures, translating the question and schema into a well-formed SQL query. Pre-trained transformer models have improved performance as in many other NLP tasks, starting with BERT-based models \cite{sqlova-2019, bridge-2020} and up to larger LLMs, such as T5 \cite{2020t5} in \cite{Scholak2021:PICARD}, OpenAPI \textit{CodeX} \cite{codex-2021} and GPT variants \cite{openai2023gpt4} in \cite{rajkumar2022evaluating, liu2023divide, pourreza2023dinsql}. 

In addition to pre-trained transformer models, several task-specific improvements have been introduced. For example, the encoding of the schema can be improved through effective representation learning, and the attention mechanism of the sequence-to-sequence model can be fine-tuned. On the decoding side, techniques that incorporate the syntactic structure of the SQL output have been proposed. These techniques can be combined in a modular and synergistic manner to further improve text-to-SQL parsing performance.

\subsection{Encoding-side: Schema Linking and Schema Encoding}

In the text-to-SQL task, the database schema is a crucial component of the input and, therefore, must be encoded to condition the generation of the SQL query along with the question. To tackle this challenge, \citet{gnn-sql-2019} proposed a graph-based approach for representing the schema, where schema items such as tables and columns are connected through association links, such as "column belongs to table" or "column is a foreign key associated with a table". They then applied a Graph Neural Network (GNN) model to learn a representation of this graph and demonstrated improved performance.

In a different approach, \citet{ratsql-2020} adopted the technique of a \textit{relation-aware self-attention mechanism} to improve the encoding of the input. This mechanism encourages the text encoder, when scanning the question, and the schema encoder, when scanning the schema, to refer to each other through a specialized self-attention mechanism. By combining this technique with BERT pre-trained contextual embeddings, the authors achieved improved schema-linking performance on unseen schema and enhanced generalization capabilities.

These results indicate the importance of properly encoding task-specific schema information and of detecting spans in the question that refer to schema items, i.e., performing schema linking.

\subsection{Decoding-side: Syntax-guided Methods}

To make sure that models generate a sequence of tokens that obey SQL syntax, different approaches have been proposed: \citet{ast-generation-2017} introduced a method where instead of generating a sequence of tokens, code-oriented models generate tokens representing the abstract syntax tree (AST) of expressions in the target program language. This strategy was adopted for text-to-SQL for example in \cite{irnet-2019, ratsql-2020}.

\citet{Scholak2021:PICARD} defined the \textit{constrained decoding method} with PICARD. PICARD is an independent module on top of a text-to-text auto-regressive model that uses an incremental parser to constrain the generated output to adhere to the target SQL grammar. Not only does this eliminate almost entirely invalid SQL queries, but the parser is also schema-aware, thus reducing the number of semantically incorrect queries as well, e.g., selecting a non-existent column from a specific table. At the time of writing, four entries in the \textit{Spider} leader board's top-ten (in terms of execution accuracy) use PICARD. Constrained decoding is a general strategy that can be easily adopted on top of any auto-regressive LLM.  We have adopted it in our approach, by designing an incremental parser for QPL, and enforcing the generation of syntactically valid query plans.

\subsection{Zero-shot and Few-shot LLM Methods}

With recent LLM progress, the multi-task capabilities of LLMs have been tested on a wide range of existing NLP tasks, including text-to-SQL. 

In zero-shot mode, a task-specific prompt is prefixed to a textual encoding of the schema and the question, and the LLM outputs an SQL query. In an evaluation of the OpenAPI Codex language model \cite{rajkumar2022evaluating, liu2023comprehensive}, it was shown that without any fine-tuning (which is required for constrained decoding, as one needs access to the scores output by beam search) Codex achieves 67\% execution accuracy, compared to the then-SOTA 79.1\% achieved by T5-3B with PICARD. In our own evaluation, we estimate that as of May 2023, GPT-4 achieves about 74\% execution accuracy under the same zero-shot prompting conditions.  Zero-shot performance is impressive (given that "no work" is required to achieve it on this task) but still lags behind much smaller specialized models such as \textit{RESDSQL-3B} \cite{resdsql-2023}.

More sophisticated few-shot LLM prompting strategies have also been investigated: in  \cite{guo2023casebased}, given an input $(Question, Schema)$, a small set of related examples (about 2 to 5) $(Question, Schema, SQL)$ is selected from the training set, and are used as prefix to the case at hand to provide a few-shot prompt. This is currently achieving state-of-the-art performance on \textit{Spider} using OpenAPI GPT4 when using a strong case retrieval method to form the few-shot prompt.

\citet{pourreza2023dinsql} also currently present one of the top performers on \textit{Spider} with DIN-SQL, which uses a complex prompting strategy over Codex and GPT-4. The approach consists of passing the input into a chain of different prompts, and collecting the output of each prompt when continuing to the next. The chain consists of a first \textit{schema linking prompt}, which asks the LLM to identify which tables and columns should be mentioned in the query; it then passes this list of schema items to a \textit{query classification prompt}, which determines the overall structure of the query (join, nested queries, intersection, union); finally it passes the list of schema items and the query structure label to a third \textit{query generation} module. Depending on the structure, this stage can involve an embedded chain, where each component of a complex query is passed to a separate prompt, yielding separate SQL components, which are then merged into a complex query. A final \textit{self-correction} prompt is used to verify that the candidate query is syntactically correct. \citet{liu2023divide} adopt a similar decomposition-based prompting approach around the syntactic structure of the expected SQL query.

Both the few-shot prompting methods close the gap and even outperform specialized text-to-SQL models (with about 85\% execution match vs. 80\% for 3B parameters specialized models on the \textit{Spider} test set), without requiring any form of fine-tuning or training for the task. We report in our approach similar benefits when generating QPL using advanced prompting. In this paper, we focus on the hardest cases of queries, which remain challenging both in SQL and in QPL.

\subsection{Simplified Target Language and Compositional Methods}

Most text-to-SQL systems suffer from a severe drop in performance for complex queries, as reported for example in DIN-SQL results where the drop in execution accuracy between simple queries and hard queries is from about 85\% to 55\% (see also \cite{lee-2019-clause}).  Other methods have been used to demonstrate that current sequence to sequence methods suffer at \textit{compositional generalization}, that is, systems trained on simple queries, fail to generate complex queries, even though they know how to generate the components of the complex query.  This weakness is diagnosed by using challenging \textit{compositional splits} \cite{compositional-splits-2020, shaw-etal-2021-compositional, gan-etal-2022-measuring} over the training data.

One of the reasons for such failure to generalize to complex queries relates to the gap between the syntactic structure of natural language questions and the target SQL queries. This has motivated a thread of work attempting to generate simplified or more generalizable logical forms than executable SQL queries. These attempts are motivated by empirical results on other semantic parsing formalisms that showed that adequate syntax of the logical form can make learning more successful \cite{guo-etal-2020-benchmarking-meaning, herzig-berant-2021-span}. 

Most notable attempts include SyntaxSQLNet \cite{yu-etal-2018-syntaxsqlnet}, IRNet with SemQL \cite{irnet-2019} and NatSQL \cite{natsql-2021}. IRNet uses a simplified form of SQL called SemQL that is used as an intermediate representation from which the eventual SQL query is inferred in an SQL-specific manner.  SemQL, however, is not a fully compositional language: it identifies a set of six distinct query structures (e.g., \textit{select superlative filter} and \textit{select order}) and provides a distinct syntax for each one that directly reflects the intent of the query.  It does not support directly arbitrarily nested sub-queries and joins (except for the less frequent union and intersection cases). It is sufficient to cover most of Spider queries, but it would not be capable to cover more complex queries such as those collected in SEDE \cite{sede-2021} which focuses on complex queries on a specific schema.

\citet{natsql-2021} and \cite{gan-etal-2022-measuring} propose another simplified intermediary target language called \textit{NatSQL} aimed at reducing the gap between questions and queries while preserving a high coverage of SQL queries. NatSQL removes the usage of keywords such as GROUP BY, JOIN, HAVING, WHERE, INTERSECT and UNION and the usage of sub-queries. It leads to a much shorter version of queries, with less repetition of schema items when compared to the original SQL syntax. The driving design force of NatSQL and the previous SemQL or SyntaxNet methodes is to make the structure of queries as similar as possible to that of natural language.  The translation of executable SQL from the inferred intermediary language uses an algorithm which relies on the properties of the database schema, in particular, the declaration of foreign keys.

Our work is directly related to this thread. Our approach in designing \textit{QPL} is different from that of NatSQL and previous intermediary languages, in that we do not attempt to mimic the syntax of natural language and create succinct programs, but instead, we aim at using a compositional semantically regular query language, where all the nodes are executable nodes which feed into other nodes in a dataflow graph.  Our method does not aim at simplifying the mapping of a single question to a whole query, but instead at decomposing a question into simpler questions, which can then be mapped to simple queries.

\section{Method: Question Decomposition and Query Planning}

Our proposed method builds on the fact that a single SQL query is not a compositional programming language, \textit{i.e.}, in most cases, it is not possible to take a part of a SQL query and get an executable, or even a syntactically valid, sub-query.
Questions intents, in contrast, can be arbitrarily complex. In our analysis of the \textit{Spider} complex data, we found queries that require as many as 11 executable steps to complete. Previous work on intermediate languages attempted to simplify the SQL queries to make it as similar to natural language syntax. In our strategy, we aim instead to decompose the intention underlying the question into a complex plan composing simple procedural data retrieval actions, that can be then combined in an arbitrary manner using a planning mechanism.

Our approach is inspired by the thread of works attempting to solve complex QA and semantic parsing using a \textit{question decomposition strategy} \cite{unsupervised-question-decomp-emnlp-2020,
decomposing-complex-q-emnlp-2021,
sparqling-emnlp-2021,
wolfson-etal-2022-weakly,
seqzero-naacl-2022,
amr-q-decomp-ijcai2022,
zhao-etal-2022-compositional,
synthetic-nq-decomposition-tacl-2023}. In this approach, the natural language question is decomposed into a chain of sub-steps. In the context of LLMs, these methods are known as \textit{chain of thought}  \cite{COT_NEURIPS2022, program-of-thoughts-2022, planandsolve-acl2023}.  

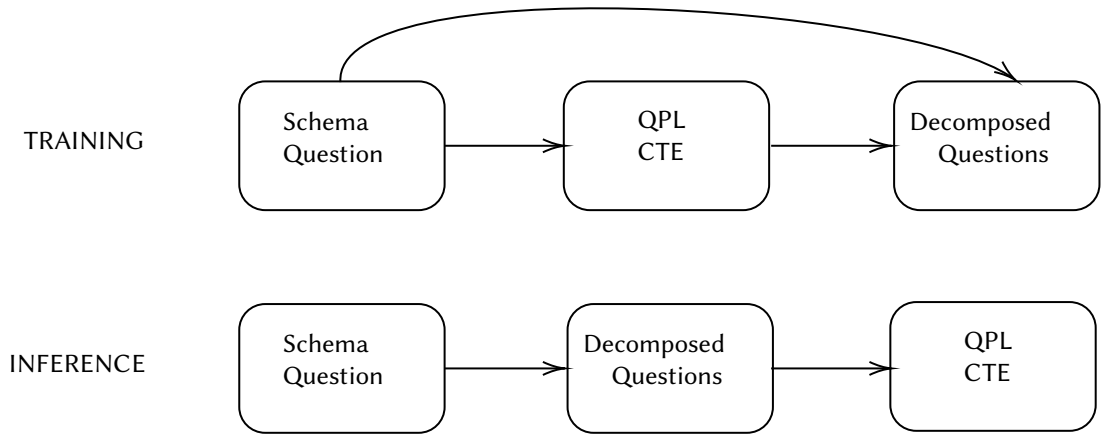
\begin{figure*}[ht]
\centering
\tikzset{every picture/.style={line width=0.75pt}} %set default line width to 0.75pt        

\begin{tikzpicture}[x=0.75pt,y=0.75pt,yscale=-1,xscale=1]

%Rounded Rect [id:dp19267676695220093] 
\draw   (146.5,100.07) .. controls (146.5,92.92) and (152.3,87.12) .. (159.45,87.12) -- (236.05,87.12) .. controls (243.2,87.12) and (249,92.92) .. (249,100.07) -- (249,138.93) .. controls (249,146.08) and (243.2,151.88) .. (236.05,151.88) -- (159.45,151.88) .. controls (152.3,151.88) and (146.5,146.08) .. (146.5,138.93) -- cycle ;

%Rounded Rect [id:dp1601907228900492] 
\draw   (308.5,100.07) .. controls (308.5,92.92) and (314.3,87.12) .. (321.45,87.12) -- (398.05,87.12) .. controls (405.2,87.12) and (411,92.92) .. (411,100.07) -- (411,138.93) .. controls (411,146.08) and (405.2,151.88) .. (398.05,151.88) -- (321.45,151.88) .. controls (314.3,151.88) and (308.5,146.08) .. (308.5,138.93) -- cycle ;

%Rounded Rect [id:dp9466791504309542] 
\draw   (473.38,100.07) .. controls (473.38,92.92) and (479.17,87.12) .. (486.33,87.12) -- (562.92,87.12) .. controls (570.08,87.12) and (575.88,92.92) .. (575.88,100.07) -- (575.88,138.93) .. controls (575.88,146.08) and (570.08,151.88) .. (562.92,151.88) -- (486.33,151.88) .. controls (479.17,151.88) and (473.38,146.08) .. (473.38,138.93) -- cycle ;

%Straight Lines [id:da3428557730129842] 
\draw    (249,119.5) -- (307,119.5) ;
\draw [shift={(309,119.5)}, rotate = 180] [color={rgb, 255:red, 0; green, 0; blue, 0 }  ][line width=0.75]    (10.93,-3.29) .. controls (6.95,-1.4) and (3.31,-0.3) .. (0,0) .. controls (3.31,0.3) and (6.95,1.4) .. (10.93,3.29)   ;
%Straight Lines [id:da6683219617759912] 
\draw    (412,119.5) -- (470,119.5) ;
\draw [shift={(472,119.5)}, rotate = 180] [color={rgb, 255:red, 0; green, 0; blue, 0 }  ][line width=0.75]    (10.93,-3.29) .. controls (6.95,-1.4) and (3.31,-0.3) .. (0,0) .. controls (3.31,0.3) and (6.95,1.4) .. (10.93,3.29)   ;
%Rounded Rect [id:dp8216875170110369] 
\draw   (146.5,212.07) .. controls (146.5,204.92) and (152.3,199.12) .. (159.45,199.12) -- (236.05,199.12) .. controls (243.2,199.12) and (249,204.92) .. (249,212.07) -- (249,250.93) .. controls (249,258.08) and (243.2,263.88) .. (236.05,263.88) -- (159.45,263.88) .. controls (152.3,263.88) and (146.5,258.08) .. (146.5,250.93) -- cycle ;

%Rounded Rect [id:dp808981945515957] 
\draw   (471.5,211.07) .. controls (471.5,203.92) and (477.3,198.12) .. (484.45,198.12) -- (561.05,198.12) .. controls (568.2,198.12) and (574,203.92) .. (574,211.07) -- (574,249.93) .. controls (574,257.08) and (568.2,262.88) .. (561.05,262.88) -- (484.45,262.88) .. controls (477.3,262.88) and (471.5,257.08) .. (471.5,249.93) -- cycle ;

%Rounded Rect [id:dp7774260619194078] 
\draw   (310.38,212.07) .. controls (310.38,204.92) and (316.17,199.12) .. (323.33,199.12) -- (399.92,199.12) .. controls (407.08,199.12) and (412.88,204.92) .. (412.88,212.07) -- (412.88,250.93) .. controls (412.88,258.08) and (407.08,263.88) .. (399.92,263.88) -- (323.33,263.88) .. controls (316.17,263.88) and (310.38,258.08) .. (310.38,250.93) -- cycle ;

%Straight Lines [id:da04779174354694371] 
\draw    (249,231.5) -- (307,231.5) ;
\draw [shift={(309,231.5)}, rotate = 180] [color={rgb, 255:red, 0; green, 0; blue, 0 }  ][line width=0.75]    (10.93,-3.29) .. controls (6.95,-1.4) and (3.31,-0.3) .. (0,0) .. controls (3.31,0.3) and (6.95,1.4) .. (10.93,3.29)   ;
%Straight Lines [id:da35321064977282535] 
\draw    (412,231.5) -- (470,231.5) ;
\draw [shift={(472,231.5)}, rotate = 180] [color={rgb, 255:red, 0; green, 0; blue, 0 }  ][line width=0.75]    (10.93,-3.29) .. controls (6.95,-1.4) and (3.31,-0.3) .. (0,0) .. controls (3.31,0.3) and (6.95,1.4) .. (10.93,3.29)   ;
%Curve Lines [id:da6091103847228352] 
\draw    (197,87.5) .. controls (193.04,30.08) and (478.21,47.15) .. (532.43,85.34) ;
\draw [shift={(534,86.5)}, rotate = 217.95] [color={rgb, 255:red, 0; green, 0; blue, 0 }  ][line width=0.75]    (10.93,-3.29) .. controls (6.95,-1.4) and (3.31,-0.3) .. (0,0) .. controls (3.31,0.3) and (6.95,1.4) .. (10.93,3.29)   ;

% Text Node
\draw (166.75,100.5) node [anchor=north west][inner sep=0.75pt]   [align=left] {Schema \\Question};
% Text Node
\draw (343.75,100.5) node [anchor=north west][inner sep=0.75pt]   [align=left] {QPL\\CTE};
% Text Node
\draw (479.63,100.5) node [anchor=north west][inner sep=0.75pt]   [align=left] {\begin{minipage}[lt]{62.83pt}\setlength\topsep{0pt}
Decomposed
\begin{center}
Questions
\end{center}

\end{minipage}};
% Text Node
\draw (37.5,111) node [anchor=north west][inner sep=0.75pt]   [align=left] {TRAINING};
% Text Node
\draw (30,223) node [anchor=north west][inner sep=0.75pt]   [align=left] {INFERENCE};
% Text Node
\draw (316.63,212.5) node [anchor=north west][inner sep=0.75pt]   [align=left] {\begin{minipage}[lt]{62.83pt}\setlength\topsep{0pt}
Decomposed
\begin{center}
Questions
\end{center}

\end{minipage}};
% Text Node
\draw (506.75,211.5) node [anchor=north west][inner sep=0.75pt]   [align=left] {QPL\\CTE};
% Text Node
\draw (166.75,212.5) node [anchor=north west][inner sep=0.75pt]   [align=left] {Schema \\Question};

\end{tikzpicture}
\caption{Overall approach}
\label{fig:architecture}
\end{figure*}

Our main contribution is that we design QPL, a target language adapted to the compositional nature of complex retrieval cases, and experiment with different question decomposition (QD) strategies, including fine-tuned QD models and LLM-based CoT approaches.  The overall strategy we pursue is depicted in Fig.~\ref{fig:architecture}:
\begin{enumerate}
    \item We convert the existing Spider dataset from triplets \verb|(schema, question, sql)| to enriched tuples \verb|(schema, question, qpl, cte, decomposed questions)|. 
    \item We train models to decompose a question, predict a QPL plan given a list of decomposed questions or a single QPL plan by fine-tuning LLMs.
    \item We convert QPL into executable CTE-SQL expressions and measure execution match.
\end{enumerate}

\subsection{Query Plan Language Dataset Conversion}

We design \textit{Query Plan Language} (QPL) as a modular dataflow language that encodes the semantics of SQL queries. We take inspiration in our semantic transformation from SQL to QPL from the definition of the execution plans used internally by SQL optimizers (e.g., \cite{system-r-1979}), specifically starting from Microsoft SQL Server Execution Plan design \cite{sql-exec-plan-2018}. We automatically convert the original \textit{Spider} dataset into a version that includes QPL expressions for all the training and development parts of \textit{Spider}.

QPL is a hierarchical representation for execution plans. It is a tree of operations in which the leaves are \textit{table reading} nodes (\verb|Scan| nodes), and the inner nodes are either unary operations (such as \verb|Aggregate| and \verb|Filter|) or binary operations (such as \verb|Join| and \verb|Intersect|). Nodes have arguments, such as the table to scan in a \verb|Scan| node, or the join predicate of a \verb|Join| node.

An important distinction between execution plans and SQL queries is that every QPL sub-plan is a valid execution plan, which returns a stream of data tuples. For example, Fig.~\ref{fig:plan} shows an execution plan of depth 2, and has 4 possible sub-plans: the two \verb|Scan| leaves, the \verb|Join| sub-plan, and the \verb|Aggregate| sub-plan, which is the complete plan.

\begin{table}[h]
  \centering
  \begin{tabular}{ll}
    \toprule
    \textbf{Operator} & \textbf{Description} \\
    \midrule
    \textbf{Aggregate} & Aggregate a stream of tuples using a grouping criterion into a stream of groups \\
    \textbf{Except} & Compute the set difference between two streams of tuples \\
    \textbf{Filter} & Remove tuples from a stream that do not match a predicate \\
    \textbf{Intersect} & Compute the set intersection between two streams of tuples \\
    \textbf{Join} & Perform a logical join operation between two streams based on a join condition \\
    \textbf{Scan} & Scan all rows in a table with optional filtering predicate \\
    \textbf{Sort} & Sort a stream according to a sorting expression \\
    \textbf{TopSort} & Select the top-K tuples from a stream according to a sorting expression \\
    \textbf{Union} & Compute the set union between two streams of tuples \\
    \bottomrule
  \end{tabular}
  \caption{Description of Operations}
  \label{tab:qpl-operators}
\end{table}

We automatically convert  SQL queries into semantically equivalent QPL plans by reusing the execution plans produced by Microsoft SQL Server 2019 query optimizer. 
QPL is a high-level abstraction of the physical execution plan produced (which includes data and index statistics). 
In QPL syntax, we reduced the number of operators to the 9 operators listed in Table~\ref{tab:qpl-operators}. We also design the operators to be regular and \textit{context free}, i.e., all operators take as input streams of tuples and output a stream of tuples, and the output of an operator only depends on its inputs. In the syntax of QPL, we ensure that all schema items are fully qualified (\verb|name.column|). We experiment with different syntactic realizations of QPL expressions, and elected the version where the steps are listed in bottom-up order, corresponding roughly to the execution order of the steps.

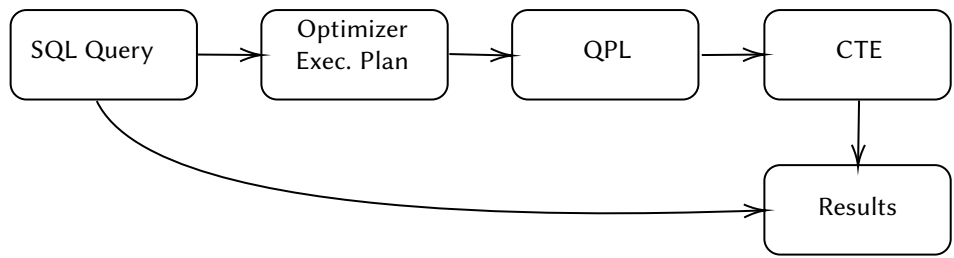
\begin{figure*}[ht]
\centering
\tikzset{every picture/.style={line width=0.75pt}} %set default line width to 0.75pt        

\begin{tikzpicture}[x=0.75pt,y=0.75pt,yscale=-1,xscale=1]

%Flowchart: Alternative Process [id:dp9970180554041281] 
\draw   (212.67,100.38) .. controls (212.67,96.03) and (216.2,92.5) .. (220.55,92.5) -- (297.8,92.5) .. controls (302.14,92.5) and (305.67,96.03) .. (305.67,100.38) -- (305.67,129.63) .. controls (305.67,133.97) and (302.14,137.5) .. (297.8,137.5) -- (220.55,137.5) .. controls (216.2,137.5) and (212.67,133.97) .. (212.67,129.63) -- cycle ;

%Flowchart: Alternative Process [id:dp7261221059703242] 
\draw   (87.5,100.38) .. controls (87.5,96.03) and (91.03,92.5) .. (95.38,92.5) -- (172.63,92.5) .. controls (176.97,92.5) and (180.5,96.03) .. (180.5,100.38) -- (180.5,129.63) .. controls (180.5,133.97) and (176.97,137.5) .. (172.63,137.5) -- (95.38,137.5) .. controls (91.03,137.5) and (87.5,133.97) .. (87.5,129.63) -- cycle ;

%Flowchart: Alternative Process [id:dp025676352707339367] 
\draw   (337.84,100.38) .. controls (337.84,96.03) and (341.37,92.5) .. (345.72,92.5) -- (422.97,92.5) .. controls (427.31,92.5) and (430.84,96.03) .. (430.84,100.38) -- (430.84,129.63) .. controls (430.84,133.97) and (427.31,137.5) .. (422.97,137.5) -- (345.72,137.5) .. controls (341.37,137.5) and (337.84,133.97) .. (337.84,129.63) -- cycle ;

%Flowchart: Alternative Process [id:dp28427288442662757] 
\draw   (463.75,100.38) .. controls (463.75,96.03) and (467.28,92.5) .. (471.63,92.5) -- (548.88,92.5) .. controls (553.22,92.5) and (556.75,96.03) .. (556.75,100.38) -- (556.75,129.63) .. controls (556.75,133.97) and (553.22,137.5) .. (548.88,137.5) -- (471.63,137.5) .. controls (467.28,137.5) and (463.75,133.97) .. (463.75,129.63) -- cycle ;

%Straight Lines [id:da8644414073146058] 
\draw    (180.5,114.88) -- (211,115.11) ;
\draw [shift={(213,115.13)}, rotate = 180.44] [color={rgb, 255:red, 0; green, 0; blue, 0 }  ][line width=0.75]    (10.93,-3.29) .. controls (6.95,-1.4) and (3.31,-0.3) .. (0,0) .. controls (3.31,0.3) and (6.95,1.4) .. (10.93,3.29)   ;
%Straight Lines [id:da8691947308240933] 
\draw    (306.5,114.63) -- (336,115.33) ;
\draw [shift={(338,115.38)}, rotate = 181.36] [color={rgb, 255:red, 0; green, 0; blue, 0 }  ][line width=0.75]    (10.93,-3.29) .. controls (6.95,-1.4) and (3.31,-0.3) .. (0,0) .. controls (3.31,0.3) and (6.95,1.4) .. (10.93,3.29)   ;
%Straight Lines [id:da5285570046398882] 
\draw    (432.5,115.13) -- (462,114.89) ;
\draw [shift={(464,114.88)}, rotate = 179.55] [color={rgb, 255:red, 0; green, 0; blue, 0 }  ][line width=0.75]    (10.93,-3.29) .. controls (6.95,-1.4) and (3.31,-0.3) .. (0,0) .. controls (3.31,0.3) and (6.95,1.4) .. (10.93,3.29)   ;

%Flowchart: Alternative Process [id:dp34312064142901644] 
\draw   (463.75,178.38) .. controls (463.75,174.03) and (467.28,170.5) .. (471.63,170.5) -- (548.88,170.5) .. controls (553.22,170.5) and (556.75,174.03) .. (556.75,178.38) -- (556.75,207.63) .. controls (556.75,211.97) and (553.22,215.5) .. (548.88,215.5) -- (471.63,215.5) .. controls (467.28,215.5) and (463.75,211.97) .. (463.75,207.63) -- cycle ;

%Straight Lines [id:da743282499836222] 
\draw    (511,138) -- (510.53,169.25) ;
\draw [shift={(510.5,171.25)}, rotate = 270.86] [color={rgb, 255:red, 0; green, 0; blue, 0 }  ][line width=0.75]    (10.93,-3.29) .. controls (6.95,-1.4) and (3.31,-0.3) .. (0,0) .. controls (3.31,0.3) and (6.95,1.4) .. (10.93,3.29)   ;
%Curve Lines [id:da6124537102234078] 
\draw    (130.5,138.25) .. controls (162.18,204.58) and (408.5,194.95) .. (462.91,193.3) ;
\draw [shift={(464.5,193.25)}, rotate = 178.32] [color={rgb, 255:red, 0; green, 0; blue, 0 }  ][line width=0.75]    (10.93,-3.29) .. controls (6.95,-1.4) and (3.31,-0.3) .. (0,0) .. controls (3.31,0.3) and (6.95,1.4) .. (10.93,3.29)   ;

% Text Node
\draw (96,106.5) node [anchor=north west][inner sep=0.75pt]   [align=left] {SQL Query};
% Text Node
\draw (222.17,96) node [anchor=north west][inner sep=0.75pt]   [align=left] {\begin{minipage}[lt]{51.49pt}\setlength\topsep{0pt}
\begin{center}
Optimizer\\Exec. Plan
\end{center}

\end{minipage}};
% Text Node
\draw (368.34,106.5) node [anchor=north west][inner sep=0.75pt]   [align=left] {\begin{minipage}[lt]{23.13pt}\setlength\topsep{0pt}
\begin{center}
QPL
\end{center}

\end{minipage}};
% Text Node
\draw (494.25,106.5) node [anchor=north west][inner sep=0.75pt]   [align=left] {\begin{minipage}[lt]{23.12pt}\setlength\topsep{0pt}
\begin{center}
CTE
\end{center}

\end{minipage}};
% Text Node
\draw (484.25,184.5) node [anchor=north west][inner sep=0.75pt]   [align=left] {\begin{minipage}[lt]{36.74pt}\setlength\topsep{0pt}
\begin{center}
Results
\end{center}

\end{minipage}};
\end{tikzpicture}
\caption{QPL generation process: the dataset SQL expressions are run through the query optimizer, which is then converted into QPL. QPL expressions are converted into modular CTE SQL programs, which can be executed. We verify that the execution results match those of the original SQL queries.}
\label{fig:qpl-cte}
\end{figure*}

In order to output executable SQL from an execution plan, we use a deterministic translation to SQL \textit{Common Table Expressions} (CTEs), also known as ``\verb|WITH| Clauses''. The translation is done bottom-up, as every clause is available to all clauses following it by the name given to the clause, similar to variable bindings in other programming languages. This produces one CTE clause per node in the QPL plan. 

The process of automatic conversion of the \textit{Spider} dataset SQL queries to QPL and their corresponding executable CTE programs is summarized in Fig.~\ref{fig:qpl-cte}. An instance of this process is shown in Fig.~\ref{fig:cte}. In the CTE translation, we obtain one expression for each node in the QPL plan. 
% In this example, we compute meaningful names for each CTE expression (e.g., \verb|Visitors_Total_Spent|).

\section{Experiments and Results}

We perform the following experiments to assess whether the compositional nature of the decompose and plan approach we have adopted succeeds at improving SQL prediction for complex questions: (1) we first attempt to directly predict QPL plans using a fine-tuned T5 LLM with Picard constrained decoding. This approach compares how learnable QPL is compared to SQL without using a question decomposition component; (2) we compare these direct QPL prediction with direct CTE predictions using a recent GPT3.5-turbo pretained LLM without question decomposition. The predicted CTEs are similar in structure to QPL, but use SQL syntax for each component. Since GPT models have been trained extensively on SQL, we determine whether a priori knowledge of SQL facilitates the generation of modular queries vs. the more abstract QPL expressions. 

\subsection{Experimental Settings}

The size of the \textit{Spider} training and development sets after QPL and CTE conversion is shown in Table~\ref{tab:sizes}. We report performance as execution match on the development set because the distribution by difficulty level is not available for the held-out Spider test set. For all fine-tuning experiments, we start from the CodeT5 Large model (770M parameters) and run fine-tuning on a single GPU using the Huggingface Transformers library. Fine-tuning converges within 2 to 3 epochs on the training data, taking about 2 hours of GPU time. For OpenAI GPT experiments, we use GPT3.5-turbo models through the OpenAI API. Experiments run for about 4 to 8 hours on throttled API calls and cost \$10 to \$30 each.

\begin{table}[h]
\centering
\begin{tabular}{crr}
\toprule
QPL Depth & Training & Development \\
\midrule
0 & 909 & 139\\
1 & 2,790 & 397 \\
2 & 1,531 & 214 \\
3 & 735 & 143 \\
4 & 117 & 30 \\
5 & 23 & 6 \\
6 & 8 & 0 \\
7 & 2 & 0 \\
\bottomrule
\end{tabular}
\caption{Size of the Spider training and development by depth of the QPL expression}
\label{tab:sizes}
\end{table}

\subsection{Baselines}

We use the following baselines for comparison: the combination of RAT-SQL encoder + NatSQL intermediary target language \cite{natsql-2021}; the development set results reported by Din-SQL with GPT4 \cite{pourreza2023dinsql}; and a direct zero-shot SQL prediction using GPT3.5-turbo (tested May 2023) which we executed.  We report the results by level of difficulty as classified in the Spider dataset in Table~\ref{tab:depth_results}. Surprisingly, GPT-based methods perform at much lower performance on the \textit{Spider} development set than on the held-out test set. As noted above, we observe a sharp drop in performance for more complex queries for all baseline models.

\begin{table}[h]
\centering
\begin{tabular}{lccc|c}
\toprule
Difficulty & NatSQL + RAT-SQL & Din-SQL GPT-4 & GPT3.5-turbo & Count\\
\midrule
Easy       & \textbf{91.6\%} & 91.1\% & 87.7\% & 227 \\
Medium     & 75.2\% & \textbf{79.8\%} & 75.1\% & 401 \\
Hard       & 65.5\% & 64.9\% & \textbf{72.5\%} & 160 \\
Extra Hard & 51.8\% & 43.4\% & \textbf{53.9\%} & 152 \\
\midrule
Overall & 73.7\% & 74.2\% & \textbf{74.3\%} & 940 \\
\bottomrule
\end{tabular}
\caption{\textit{Spider} Development Set baseline execution accuracy by difficulty level}
\label{tab:depth_results}
\end{table}

\subsection{Direct QPL and CTE Prediction}

In this experiment, we fine-tune a T5-Large LLM (770M parameters) to directly predict the QPL plan given the \verb|(schema, question)| input. We implement the PICARD constrained decoding method on the QPL syntax and at inference time, use a beam-search of 8 beams and a context of 8 tokens for partial parsing. 

\begin{table}[h]
\centering
\begin{tabular}{c|cc||cc}
\toprule
\textbf{Gold QPL Depth} & \textbf{Direct QPL} & \textbf{Direct CTE} & \textbf{Difficulty} & \textbf{Direct QPL}\\
\midrule
0 & 87.1\% & 77.7\% & easy      & 81.9\% \\
1 & 75.5\% & 69.8\% & medium    & 70.1\% \\
2 & 65.9\% & 60.3\% & hard      & 36.9\% \\
3 & 72.4\% & 49.7\% & extra     & 48.0\% \\
4 & 63.2\% & 30.0\% & & \\
\midrule
Overall & \textbf{73.6\%} & 64.4\% & & 63.4\% \\
\bottomrule
\end{tabular}

\caption{Execution match accuracy by depth of query plan and difficulty level for direct QPL using a fine-tuned T5-large and direct CTE prediction with few-shot GPT-3.5-turbo.}
\label{tab:depth_results}
\end{table}

About 8\% of the samples in the development set produce valid QPL plans but our conversion procedure fails to convert them to valid executable CTEs (these errors are partly QPL prediction errors and mostly deterministic conversion issues which we are currently processing). When considering these as failures, we obtain an overall execution match performance of 63.4\%.  When considering only executable predicted queries (which is an assessment of the confidence of the predictor) we obtain about 73.6\%. 

In Table~\ref{tab:depth_results}, we use two distinct metrics to classify queries by difficulty level: on the right, the \verb|easy| to \verb|extra-hard| labels provided in the original Spider dataset. These labels are based on a syntactic analysis of the original SQL query (e.g., whether it contains a group-by statement). The QPL metric we use to assess complexity is the natural depth of the QPL tree.  We have verified that these two metrics are not well correlated. We also observe that the fine-tuned models have a natural decrease of performance for deeper QPL trees, whereas the relation between performance and Spider difficulty label is more chaotic.

The overall performance is similar to that obtained by state of the art Spider systems (74.3\% for GPT3.5-turbo without fine-tuning). We observe, however, improved performance on difficult queries. (63.4\% average for hard/extra-hard with GPT vs. 70.8\% average for QPL plans with height 3 and 4). This result indicates that learning QPL with a fine-tuned small model is competitive with the largest existing models and outperforms them on complex queries, demonstrating the benefit of a compositional intermediary language.  

Table~\ref{tab:root} indicates how well the fine-tuned model succeeds at predicting the root operator of QPL plans. We identify that the model suffers on JOIN operations, which indicates that a decomposition-based approach has potential benefits.

\begin{table}[h]
\centering
\begin{tabular}{l|ccc|r}
\toprule
 & Precision & Recall & F1 & Support \\ 
\midrule
Aggregate & 0.89 & 0.95 & 0.92 & 251 \\
Except & 0.93 & 0.93 & 0.93 & 56 \\
Filter & 0.94 & 0.95 & 0.95 & 66 \\
Intersect & 0.84 & 0.96 & 0.90 & 27 \\
Join & 0.82 & 0.63 & 0.71 & 149 \\
Scan & 0.78 & 0.88 & 0.83 & 139 \\
Sort & 0.84 & 0.76 & 0.80 & 83 \\
TopSort & 0.91 & 0.96 & 0.93 & 159 \\
Union & 1.00 & 0.25 & 0.40 & 4 \\
\midrule
Accuracy & & & 0.87 & 934 \\
\bottomrule
\end{tabular}
\caption{Root-operator prediction performance metrics per operator type.}
\label{tab:root}
\end{table}

We compare direct QPL prediction with direct prediction of CTE SQL expressions using a few-shot GPT3.5-turbo prompting method. The results are shown in the second column of Table~\ref{tab:depth_results}. They are overall lower than the fine-tuned T5-large model trained on direct QPL prediction, confirming that the abstract syntax of QPL is easier to learn than SQL syntax, even on single-clause SQL statements present in CTE programs.

% used OpenAI GPT3.5-turbo with this part added to the prompt:
% @Michael - should we add the prompt I used for the baseline or perhaps write the full prompt here?

%\begin{minted}{text}
%Use SQL CTE clauses only, and multiple CTE clauses with only simple SQL queries and no embedded sub-queries at all, so that the code will be modular, in the following style:
%\end{minted}
%\begin{minted}{sql}
%WITH CTE_Has_Population AS ( SELECT City_ID FROM city WHERE Population > 0 ), CTE_Farm_Competition AS ( SELECT Theme, Host_city_ID FROM farm_competition ), CTE_Theme AS ( SELECT Theme FROM CTE_Has_Population JOIN CTE_Farm_Competition ON CTE_Has_Population.City_ID = CTE_Farm_Competition.Host_city_ID ) SELECT * FROM CTE_Theme
%\end{minted}

\section{Conclusion and Future Work}

We address the task of text-to-SQL as an instance of semantic parsing, mapping questions to QPL, a compositional planning language for data retrieval operations. We designed QPL in such a way that it has high-coverage (over 99\% of Spider's SQL queries can be automatically converted to QPL expressions) and can be converted to modular executable SQL programs with CTEs. 

Initial experiments demonstrate that modern fine-tuned LLMs or prompt-based LLM models succeed to generate QPL expressions overall as well as SQL queries. QPL prediction, however, performs better on highly complex query plans than state-of-the-art text-to-SQL systems. This is particularly important in the context of interactive data-exploration assistants for non-programming users. QPL provides an interpretable data exploration layer which can be verified by users, as opposed to highly complex SQL queries.

In future work, we study how to exploit the modularity of QPL plans, by decomposing original questions in the data into high-level plans, which can then be translated step by step into full QPL expressions, as shown in Fig~\ref{fig:decomp}. To this end, we must learn how to decompose a question into a sequence of simpler questions.

\begin{figure*}[ht]
\raggedright
\textbf{Question}: What is the total ticket expense of the visitors whose membership level is 1?

\textbf{\\Decomposed Questions:}
To answer the question: \textit{What is the total ticket expense of the visitors whose membership level is 1?}, answer the following sub-questions:
\begin{enumerate}
    \item Who are the visitors with membership level 1?
    \item What is the total spent by visitors during their visits?
    \item What is the total spent by each visitor with membership level 1 during their visits?
    \item What is the total spent by all visitors with membership level 1 during their visit?
\end{enumerate}

    \caption{Original question and the derived decomposed questions.}
    \label{fig:decomp}
\end{figure*}

There is no available training data that pairs complex questions with decomposed questions for text-to-SQL in a supervised manner. We, therefore, consider three alternative approaches for learning question decomposition:
\begin{itemize}
    \item We train a \textit{reverse parser} on the Spider-QPL data, which predicts how to formulate a question given a QPL plan. We fine-tune a T5-large model (770M parameters) on the dataset for this task of question prediction from QPL.
    \item We use a few-shot prompting approach on OpenAI GPT3.5-turbo (which probably contains about 200Bn parameters).
    \item We use the QMDR question decomposition model introduced in \cite{qdmr-tacl-2020, wolfson-etal-2022-weakly}.
\end{itemize}

The planning task that we finally investigate consists of taking into account the question decomposition to condition QPL generation.

\section*{Acknowledgments}

We thank the Frankel Center for Computer Science for their support.

%%
%% Define the bibliography file to be used
\bibliography{ep}

% \appendix

\end{document}